\documentclass{article}

\usepackage{arxiv}

\usepackage[utf8]{inputenc} 
\usepackage[T1]{fontenc}    
\usepackage{hyperref}       
\usepackage{url}            
\usepackage{booktabs}       
\usepackage{amsfonts}       
\usepackage{nicefrac}       
\usepackage{microtype}      
\usepackage{cleveref}       
\usepackage{lipsum}         
\usepackage{graphicx}
\usepackage{natbib}
\usepackage{doi}

\usepackage{here}

\title{Proposing Novel Extrapolative Compounds by Nested Variational Autoencoders}

\author{
	\href{https://orcid.org/0000-0001-7217-6097}{\includegraphics[scale=0.06]{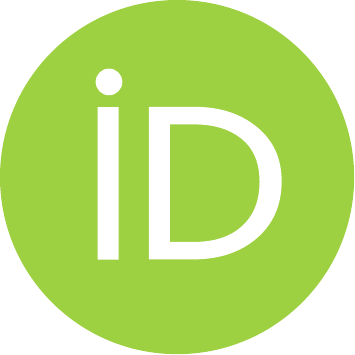}\hspace{1mm}Yoshihiro Osakabe}\\
	Research and Development Group\\
	Hitachi, Ltd.\\
	Tokyo, Japan 1858601 \\
	\texttt{yoshihiro.osakabe.fj@hitachi.com} \\
	\And
	\href{https://orcid.org/0000-0002-4727-1073}{\includegraphics[scale=0.06]{imgs/orcid.pdf}\hspace{1mm}Akinori Asahara}\\
	Research and Development Group\\
	Hitachi, Ltd.\\
	Tokyo, Japan 1858601 \\
	\texttt{akinori.asahara.bq@hitachi.com} \\
}

\renewcommand{\shorttitle}{Proposing Novel Extrapolative Compounds by Nested VAEs}

\hypersetup{
pdftitle={Proposing Novel Extrapolative Compounds by Nested Variational Autoencoders},
pdfauthor={Y. Osakabe, A. Asahara},
pdfkeywords={Automatic molecular generation, Deep generative model, Materials Informatics, Variational autoencoder},
}

\begin{document}
\maketitle

\begin{abstract}
	Materials informatics (MI), which uses artificial intelligence and data analysis techniques to improve the efficiency of materials development, is attracting increasing interest from industry. One of its main applications is the rapid development of new high-performance compounds. Recently, several deep generative models have been proposed to suggest candidate compounds that are expected to satisfy the desired performance. However, they usually have the problem of requiring a large amount of experimental datasets for training to achieve sufficient accuracy. In actual cases, it is often possible to accumulate only about 1000 experimental data at most. Therefore, the authors proposed a deep generative model with nested two variational autoencoders (VAEs). The outer VAE learns the structural features of compounds using large-scale public data, while the inner VAE learns the relationship between the latent variables of the outer VAE and the properties from small-scale experimental data. To generate high performance compounds beyond the range of the training data, the authors also proposed a loss function that amplifies the correlation between a component of latent variables of the inner VAE and material properties. The results indicated that this loss function contributes to improve the probability of generating high-performance candidates. Furthermore, as a result of verification test with an actual customer in chemical industry, it was confirmed that the proposed method is effective in reducing the number of experiments to $1/4$ compared to a conventional method.
\end{abstract}


\section{Introduction} 
\label{sec:introduction}

Technological innovation in the materials industry is important for a wide range of related manufacturing industries, and materials informatics (MI) that utilizes AI and data analytics technology has attracted much interest due to recent advances. In particular, MI is expected to make it possible to find new materials that meet the required performance in a shorter period of time than before by utilizing past experimental data and numerical simulation data.

For example, in the development of organic materials, the virtual screening method is used to efficiently extract candidate compounds to be tested. In this method, a machine learning model trained with experimental data is used to screen a large number of chemical formulas by predicting their properties. If the prediction accuracy is high, unnecessary experiments can be avoided and the number of experiments can be reduced. However, there are two problems with this method.

The first problem is the extrapolability of the prediction. Machine learning models commonly used in virtual screening methods are interpolative, which can only make effective predictions within the range of the training data. However, the task of searching for new materials that outperform known materials is a matter of extrapolability. As a result, promising compounds that were initially expected to be discovered may be sifted out.

The second problem is the quality of the screening targets. When screening a list of compounds irrespective of their performance, it is difficult to shorten the time to discover new materials even if the prediction accuracy is high. Conventionally, the "structure generation process," which prepares compounds to be screened, has been the bottleneck. For example, BRICS\citep{Degen2008brics} method, which generates structures by mechanically combining substructures of known compounds, is inefficient because it generates a large number of chemical structures that cannot exist in nature.

Recently, various deep generative models have been proposed to more directly generate compounds with high probability of performance improvement. Gómez-Bombarelli~et~al.~proposed ChemicalVAE~\citep{Gomez-Bombarelli2018}, which uses a variational autoencoder (VAE) to obtain a continuous representation from a chemical formula expression SMILES~\citep{Weininger1988} given in string form. In general, the latent representation of a VAE is represented by a real-valued vector, which is called a latent vector. ChemicalVAE has a separate neural network that predicts the properties of compounds from the latent vectors of VAE, and the output of this predictor is also used for training the VAE. With this feature, the latent space of ChemicalVAE reflects both the similarity of chemical structures and the similarity of property values.
In addition to being able to generate new chemical formulas by specifying arbitrary latent vectors, ChemicalVAE has been shown to be able to discover compounds with optimal properties by exploring the latent space. However, deep generative models such as VAE commonly require a huge amount of training data to achieve sufficient performance. Practically, it is difficult to collect a large amount of experimental data in which chemical formulas are paired with property values. In fact, it is often possible to accumulate only a hundred to a thousand data at most in practical cases, and a method that can be effective even with a small amount of experimental data.

Therefore, Osakabe~et~al.~proposed a deep generative model, called MatVAE, which consists of two nested VAEs independently trained on different datasets~\citep{osakabe2021aaaimlps}. The first (outer) VAE, which is trained on a huge open dataset of SMILES, is a universal generator of chemical structural formulae. The second (inner) VAE, which is trained on a small experimental dataset, learns the structure–property relation.
This nested structure allows us to generate candidate compounds with high probability of having the desired property values.
Their previous work indicates that the model can generate more than five times as many compounds with improved property compared to ChamicalVAE in the case of the experimental data is limited to 1000 records~\citep{osakabe2021aaaimlps}.
Their network structure and the learning method are designed to perform even with small amounts of experimental data. However, in the generation process, MatVAE are only searching the latent space for the neighborhood of the best performing compound among the known compounds.
For further improvement in efficiency, more direct and extrapolative approaches are required to generate candidates with more desirable performance.

In this study, we propose a method that can generate high-performance compounds more directly than nearest neighbor search in latent space.
Specifically, we introduce a loss function into MatVAE such that a given component of the latent vector is strongly correlated with a property value, thereby creating a linear dimension in the latent space with respect to the property value.
If the ideal latent space can be obtained by learning, it is expected that the properties of the generated compound can be adjusted according to the bias to the correlated component.

\section{Related works: deep generative models for chemical design} 
\label{sec:related_works}

Deep neural networks (DNNs) have outperformed other methods in various fields on regression and classification tasks. DNN have been also utilized to generate a novel sample similar to the training data, i.e., a deep generative model (DGM). DGM assumes that the observed data $x$ is generated from an unobserved latent variable $z$ and aims to learn a transformation rule $p(x|z)$.

DGM has been extensively studied to directly obtain chemical structures that have desirable properties.
Previously, methods for training a generative adversarial network (GAN) with the reinforcement learning (RL) framework have been reported~\citep{Olivecrona2017,DeCao2018}.
The major difference between them is the representation of the compounds; REINVENT~\citep{Olivecrona2017} and MolGAN~\citep{DeCao2018} use text-based and graph-based representations, respectively. 
Another approach uses variational autoencoder (VAE) models; for example, JT-VAE using graph-based representation~\citep{Jin2018} and ChemicalVAE using text-based representation~\citep{Gomez-Bombarelli2018}.
With VAEs, the chemical structure can be directly obtained by specifying one point in its latent space.
Studies have shown that it is possible to optimize the property values by searching the latent space because of the continuity of the space.
However, it is important to note that the previous studies required huge training datasets. ChemicalVAE and REINVENT were trained using supervised learning with 250,000 and 350,000 compound data extracted from the ZINC database~\citep{Irwin2012}, respectively.
In most cases, when developing industrial chemical products, only a few hundred or thousand supervised training data, i.e., data with property values measured by experiments, are available for a single product family.


\section{Proposal: MatVAE} 
\label{sec:proposal_matvae}

\subsection{Variational autoencoder}

Variational autoencoder (VAE) looks for the generative model

\begin{eqnarray}
    p(x) =  \int p(x|z; \theta) p(z) \,dx 
\end{eqnarray}

where $x$ is a point from the data set, $z$ is the vector of latent variables, and $\theta$ are model parameters.
VAE adopts the assumption that both $p(x|z; \theta)$ and $p(z)$ are normal distributions,

\begin{eqnarray}
    p(x|z; \theta) &=&  \mathcal{N}(x|f(z; \theta), \sigma^2 I), \\
    p(z) &=& \mathcal{N}(0, I)
\end{eqnarray}
and then approximates the deterministic function $f(z;\theta)$ with a neural network. To approximate the integral, VAE implements an autoencoder structure to optimize a variational bound on the data distribution $p(x)$, from which VAE learns to sample.
An input x is first transformed with an encoder network into a distribution on latent variables (modeled directly with the mean and variance), then a latent vector $z$ is sampled from that distribution (in reality, VAE samples e from $\mathcal{N}(0, I)$ and then transforms it into $z$: $\mathcal{N}(\mu(x), \Sigma(x))$ with the corresponding linear transformation), and then it is transformed with a decoder network into a reconstruction $y$ of the original vector $x$. To sample from a trained VAE, one samples the latent vector $z$ from $\mathcal{N}(0, I)$ and then transforms it into a sample with the decoder network.

\subsection{Methodological overview} 
\label{sub:methodological_overview}

In order to realize a DGM that generates chemical formulas as candidate compounds for actual experiments, VAE has to learn two objectives. The first is the characteristics of chemical structures that can actually exist. When the training data is a representation of a compound in string form, such as SMILES~\citep{Weininger1988}, the main task is to learn the pattern of elemental symbol sequences. The second is the relationship between the structural features of the compound and its properties.
It is considered that trying to acquire these two learning objectives in a single network causes the unnecessary requirement of a large amount of experimental data for training. Thus, in the proposed method, VAEs are trained independently for each of these two learning objectives.

\begin{figure}[H]
\includegraphics[width=\textwidth]{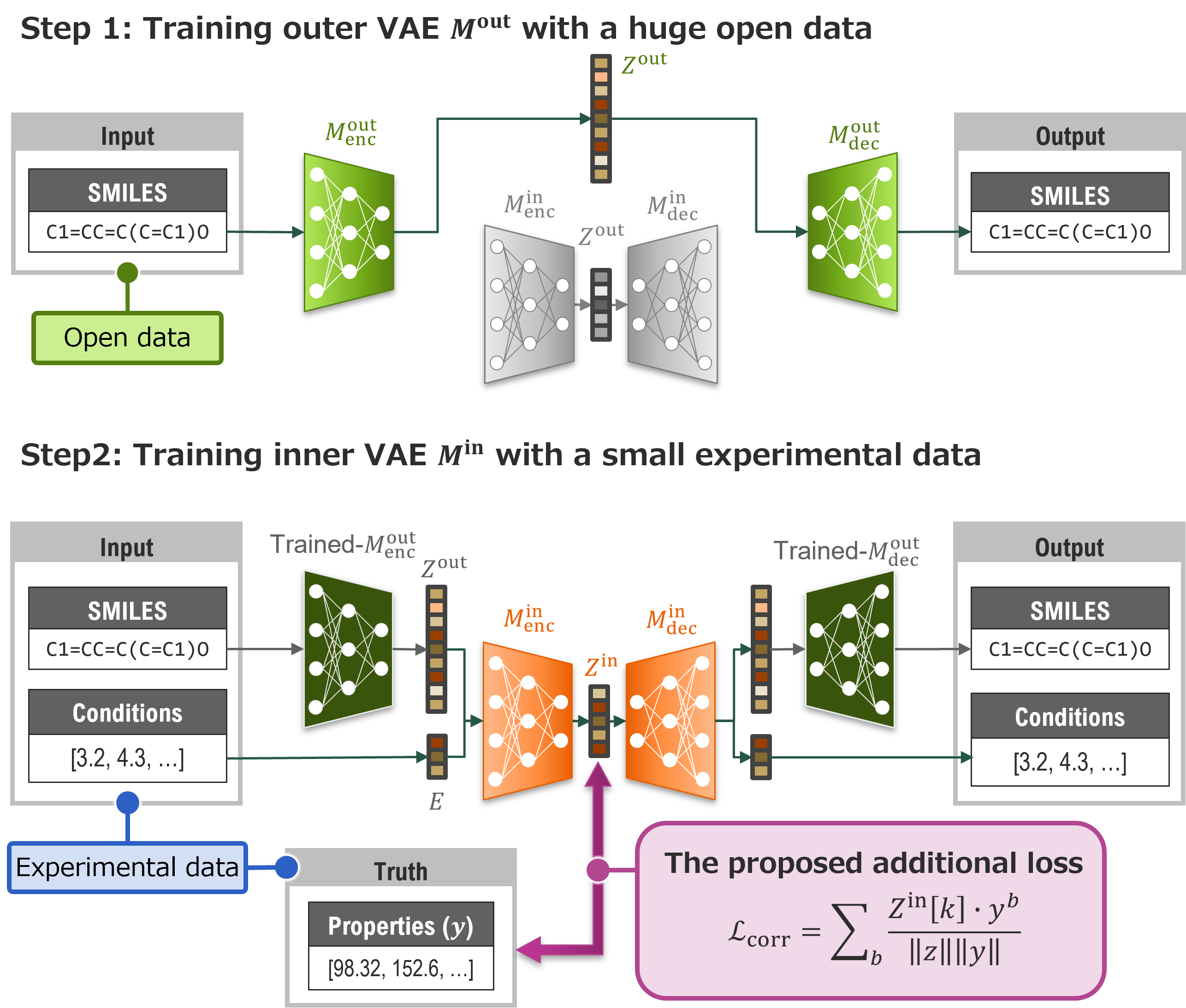}
\caption{Schematic diagram of the proposed model, MatVAE. The model consists of two VAEs, and they are trained independently in two steps.} \label{overview}
\end{figure}

Figure~\ref{overview} shows a schematic diagram of the proposed model MatVAE.
As described earlier, it consists of two VAEs.
The outer network ($M^{out}$) is a VAE that acquires structural features of compounds as latent expressions based on a large number of chemical formulas obtained from open data.
Its encoder, $M_{\rm enc}^{\rm out}$, consists of several 1D convolutional layers and several fully connected layers.

As shown in Fig. \ref{overview}, the training process is divided into two steps, and $M^{\rm out}$ is trained first.
The training data is a dataset of tens or hundreds of thousands of chemical formulae curated from open data, ZINC~\citep{Irwin2012}.
A common loss function $L_{\rm recon}$ for VAE~\citep{Kingma2013} is applied so that $M_{\rm dec}^{\rm out}$ outputs the same array $H$ that is input to $M_{\rm enc}^{\rm out}$, and the weights and other parameters of $M^{\rm out}$ are optimized.
Next, $M^{\rm in}$ is trained with about a hundred or a thousand experimental data.
As same as the training of $M^{\rm out}$, the normal VAE loss $L_{\rm recon}$ is used for training $M^{\rm in}$.
In this study, we propose an additional loss function $L_{\rm corr}$ such that a predetermined component of $Z^{\rm in}$ is strongly correlated with the property value as described in the next section.

\newpage

\subsection{Representation of compounds}

In this paper, the generation target is chemical structural formulae given as string-based format, SMILES~\cite{Weininger1988}.
The SMILES strings are tokenized into characters defined by SMILES grammar, then they are transformed into one-hot vector by following the usual mannar of natural language process.
Thus, a single SMILES representation becomes an array of $d \times M \times N$ dimensions, where $d$ is a batch size, $M$ is a number of chemical symbols, and $N$ is the max length of the SMILES string, respectively.
The reconstruction loss $\mathcal{L}_{\rm vae}$ is calculated with the input and the reconstructed one-hot vectors.
The character-by-character nature of the SMILES representation and the fragility of its internal syntax (opening and closing cycles and branches, allowed valences, etc.) can still result in the output of invalid molecules from the decoder, even with the variational constraint. When converting a molecule from a latent representation to a molecule, the decoder model samples a string from the probability distribution over characters in each position generated by its final layer. As such, multiple SMILES strings are possible from a single latent space representation. We employed the open source chem--informatics suite RDKit~\cite{rdkit} to validate the chemical structures of output molecules and discard invalid ones. While it would be more efficient to limit the autoencoder to generate only valid strings, this postprocessing step is lightweight and allows for greater flexibility in the autoencoder to learn the architecture of the SMILES.

In this study, instead of the widely used SMILES, SELFIES~\citep{Krenn2020} was adopted as the representation of the compound due to its high robustness.
SELFIES is compatible with SMILES, and was proposed for use in generative models such as VAE and GAN. Although this makes the string redundant, it enables a network very robust expression learning. For example, the following two notations both refer to the same compound and can be uniquely converted to each other.

\begin{description}
   \item[SMILES] \verb|C1=CNCC1|
   \item[SELFIES] \verb|[C][=C][N][C][C][Ring1][Branch1_1]|
\end{description}

\noindent One SELFIES representation becomes an array $H$ of $d \times M \times N$ dimensions, where $d$ is a batch size, $M$ is a number of chemical symbols, and $N$ is the max length of the SELFIES string, respectively.
Accordingly, $M_{\rm enc}^{\rm out}$ takes $H$ as its input, and outputs a $P$-dimensional latent vector, $Z^{\rm out}$.
The decoder $M_{\rm dec}^{\rm out}$ consists of several fully connected layers and Gated Recurrent Unit (GRU) layers, which inversely convert the $P$-dimensional vectors into $H$.

The inner network ($M^{\rm in}$) takes the ($P+Q$)-dimensional vector concatenated with $Z^{\rm out}$ and $Q$-dimensional vector representing experimental conditions such as the temperature, amount of compounding, and catalyst during synthesis.
The encoder $M_{\rm enc}^{\rm in}$ and the decoder $M_{\rm dec}^{\rm in}$ both consist of several fully connected layers, and its latent variable $Z^{\rm in}$ is an $R$-dimensional vector where $R<P+Q$.

\subsection{Loss Function}

The VAE loss function consists of two parts: VAE tries to make the latent variable distribution close to $\mathcal{N}(0, I)$ (minimizing KL distance between them) and at the same time makes $y$ as similar to $x$ as possible (minimizing reconstruction loss, $\mathcal{L}_{\rm vae}$).
In many cases of material development, the goal is to discover a new compound that outperforms the existing performance for a certain property.
In other words, we want to propose a compound that exists in the extrapolation region of the known data for the target property value.
Therefore, we designed the loss function $\mathcal{L}_{\rm corr}$ so that the predefined $k$-th component $z_k$ of $z$, where $k$ is an integer such that $0\leq k<R$, has a strong correlation with property values.
For instance, $\mathcal{L}_{\rm corr}$ can be defined by following the cosine similarity between the $b$-th properties ${\rm y}^{b}$ and the $z_k$, such as

\begin{eqnarray}
    \mathcal{L} &=& \mathcal{L}_{\rm vae} - \alpha \mathcal{L}_{\rm corr},\label{eq:LossAll}\\
    \mathcal{L}_{\rm corr} &=& \sum_{b} \frac{z[k] \cdot {\rm y}^{b}}{\|z\| \|y\|}.\label{eq:LossCorr}
\end{eqnarray}

\noindent
The correspondence between $b$ and $k$, in other words, which property and components are correlated, is determined beforehand.
The whole loss function $L$ for encoder is given in \ref{eq:LossAll}, where $\alpha$ is a hyperparameter that takes positive value because $\mathcal{L}_{\rm corr}$ will becomes large if the training succeeds.

The combination of the trained decoder and encoder is used as a generative model for automatic chemical design.
Due to the collaboration of the two networks, the latent space $z$ is a continuous representation in which the similarity of chemical structures and the similarity of properties are simultaneously reflected as distances.
Furthermore, due to the effect of $\mathcal{L}_{\rm corr}$, $z_k$ is correlated with the objective property.
Hence, if the ideal learning is achieved, $z_k$ can be used as a parameter for candidate generation.
In other words, if we encode a chemical formula that has particularly high performance among known compounds, then add a certain bias to its $z_k$, and finally decode the latent vector $z$, we can directly obtain a chemical formula for experimental candidates with high accuracy of performance improvement.
This method would allow us to obtain the latent vector that gives improved compounds more directly than the conventional neighborhood sampling in the latent space.

\newpage

\begin{figure}[H]
    \includegraphics[width=0.96\textwidth]{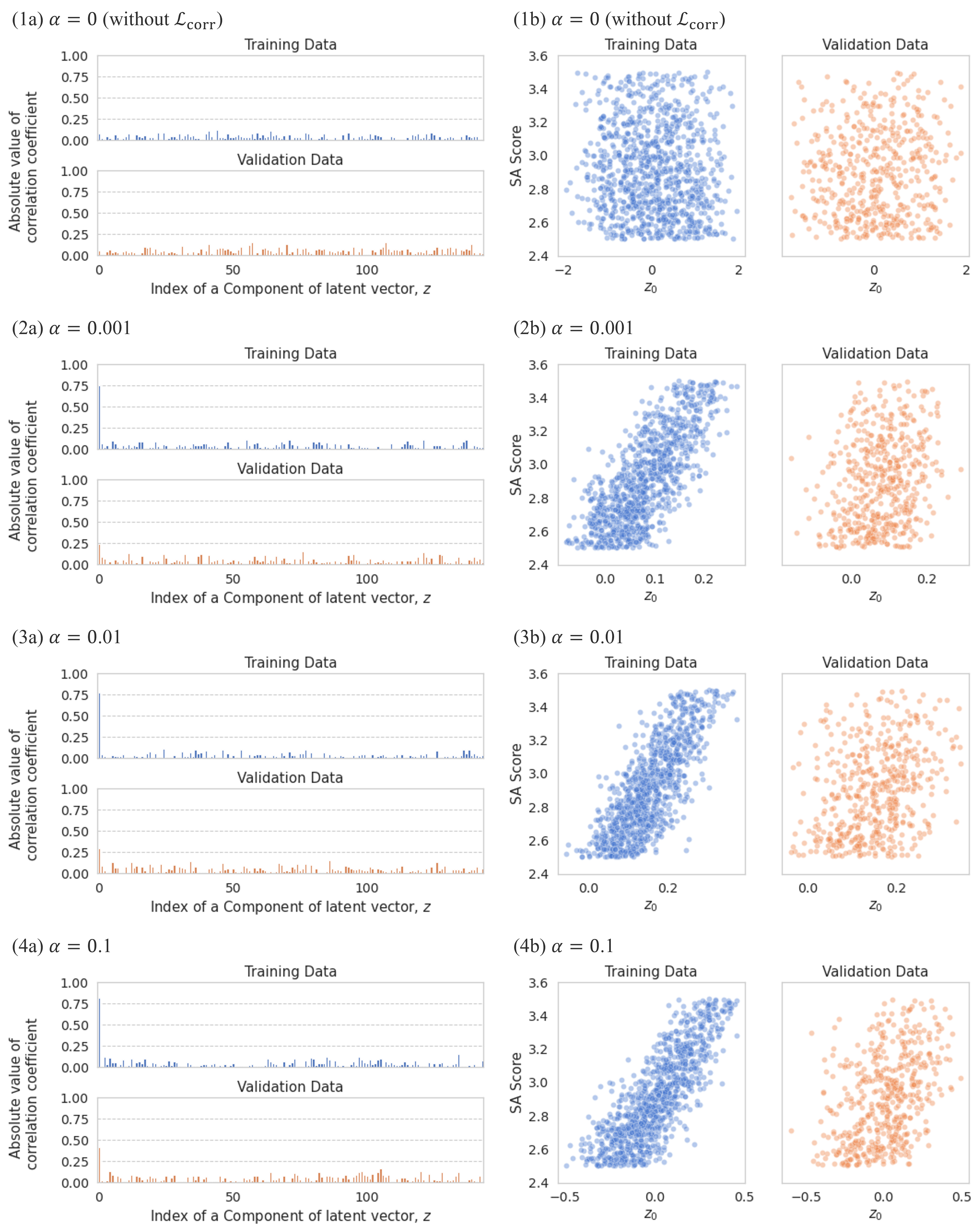}
    \centering
    \caption{
        The detailed results of training with different $\alpha$ to verify the effect of the proposed additional loss function, $\mathcal{L}_{\rm corr}$.
        (1a)--(4a) describe the absolute values of the correlation coefficients between each component of latent variables $z$ and SA Scores.
        (1b)--(4b) are scatter plots of $z_0$ and SA score.
    } \label{training_results}
\end{figure}

\newpage

\section{Experiments} 
\label{sec:experiments}

\subsection{Datasets}

To verify the effectiveness of the proposed loss function $L_{\rm corr}$, the following experiments are conducted.
In principle, the performance of the model should be evaluated through actual chemical experiments.
However, in this study, we prioritize the investigation of the informatics aspect of the proposed method and conduct experiments by treating the Synthetic Accessibility Score (SAS)~\citep{Ertl2009} as a substitute for actual properties.
SAS, $S$, is an index value that quantifies the difficulty of synthesis as a real number between $S=1$ (easy) and $S=10$ (hard), and can be calculated only from SMILES using the open source software, RDkit~\citep{rdkit}.

All the data used in the experiments were randomly selected data sets from an open database of commercial compounds called ZINC~\citep{Irwin2012}. We prepared two types of the training data, one to be considered as open data and the other as experimental data. One of the objectives of MatVAE is to generate candidate compounds that have improved property values. In other words, the model is expected to generate compounds with SA score outside the range of the given training dataset. Thus, the range of SA scores of the training dataset for inner VAE is deliberately limited as described in Table~\ref{datasets}. It means that the inner VAE cannot learn the structure–property relation in the range under 2.5 and over 4.5.

\begin{table}[h]
\centering
\caption{The details of the training datasets}\label{datasets}
\begin{tabular}{|l|l|l|l|}
\hline
Role & contains & \# of records & conditions\\ \hline
open data & SELFIES & 500,000 & None\\ \hline
experimental data & SELFIES with SAS & 1,000 & $2.5<S<4.5$\\ \hline
\end{tabular}
\end{table}

\subsection{Evaluation of the proposed loss function}

To confirm the effect of the proposed loss function $L_{\rm corr}$, we examined the correlation coefficients between each component of the latent variable $Z^{\rm in}[k]$ and SAS by calculating from 1000 compounds included in the experimental dataset.
The figure on the left,~\ref{training_results}~(1a)--(4a), plots the correlation coefficient between each component of the latent variable and the SA score. The larger the value of $alpha$, the larger the correlation coefficient for the validation data set, indicating that the learning effect is reflected.
In the figure on the right,~\ref{training_results}~(1b)--(4b), scatter plots of SA scores versus $Z^{\rm in}[0]$. In fact, we can see that the trend of the physical properties follows the value of $Z^{\rm in}[0]$, indicating that they are founded without significant outliers.

\begin{figure}[H]
    \includegraphics[width=0.5\columnwidth]{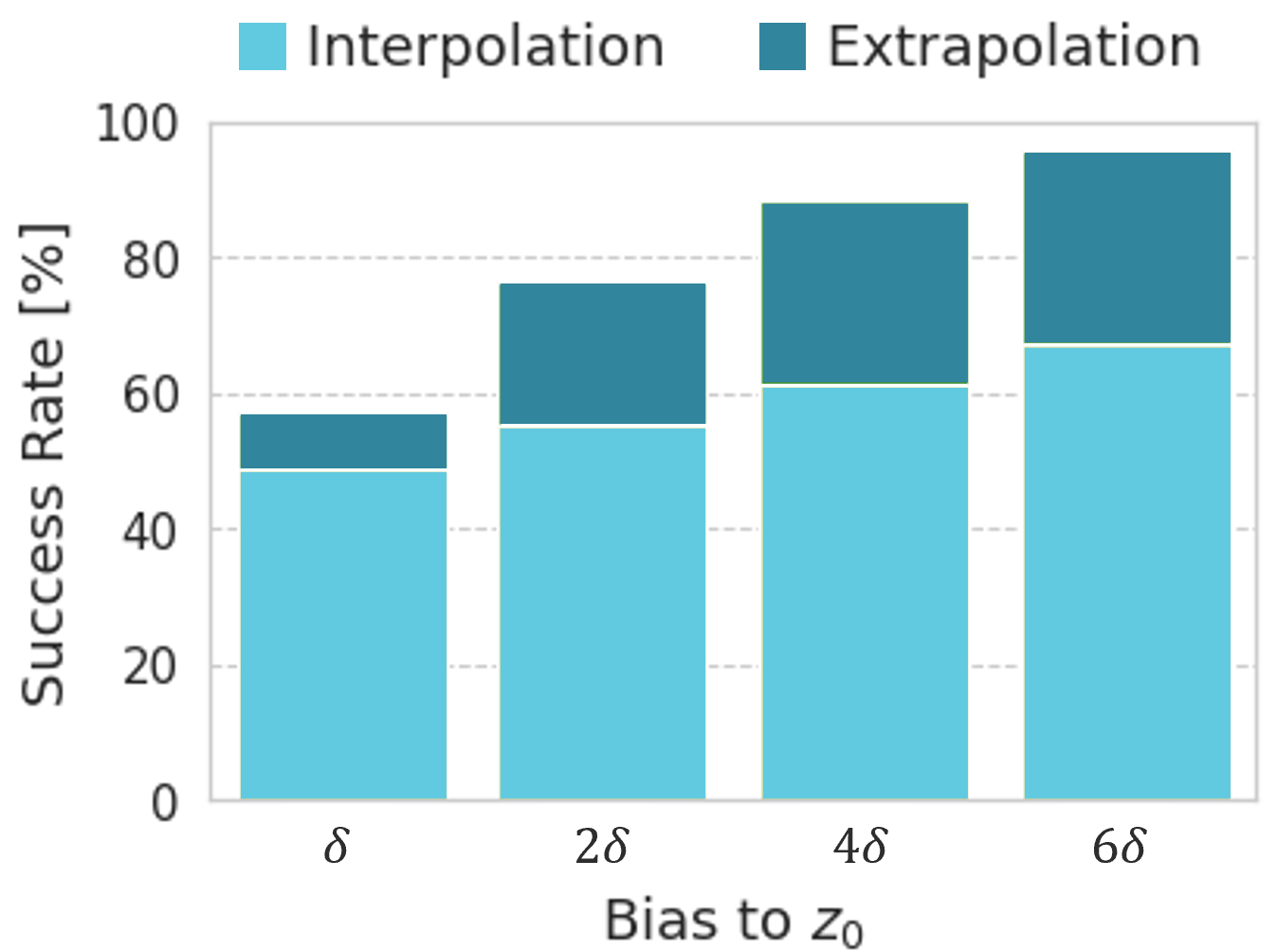}
    \centering
    \caption{
        Success rate of generated compounds with improved SA Score by adding a bias to $Z^{\rm in}[0]$.
        It also indicates the ratio of whether the SA scores of the generated compounds were interpolated or extrapolated with respect to the training data.
    } \label{rate_extrapolation}
\end{figure}

Next, we investigated whether this $Z^{\rm in}[0]$ can actually be utilized as a parameter for the generation of improved candidate compounds.
We applied positive biases to the latent vector $Z^{\rm in}[0]$ obtained by encoding a known compounds, and then calculated the SAS of the decoded SELFIES from the biased $Z^{\rm in}$.
It is considered a success when the SAS of generated compounds becomes larger than the original SAS when a positive bias is added.
Figure~\ref{rate_extrapolation} shows a plot of such success rate after 1000 trials.
It was demonstrated that the SAS of the generated candidates can be selectively modulated depending on the magnitude of the applied bias.

Finally, we investigated whether the proposed method can produce compounds that exceed the existing properties, which is the ultimate goal of material development.
The prepared experimental data was filtered so that their SAS is in the range of $2.5<S<4.5$.
Therefore, if the SAS of generated compounds exceeds this range, it can be understood that the proposed model could generate compounds that exist in the extrapolation region.
Figure~\ref{rate_extrapolation}~(b) shows the probability of generating compounds having $S>4.5$ when 1000 compounds were generated by adding a positive bias to $Z^{\rm in}[0]$. 
This task of proposing a compound with a larger SAS can be understood as a task of generating the structure of a more complex compound, which can only be successful if the relationship between structure and properties is properly acquired.
The experimental results show that the model is successful in proposing compounds with extrapolatory SAS with high probability. In fact, the model outputs longer and more diverse SELFIES to the extent that they satisfy the grammar rules.

These results shown in this section indicate that the proposed loss function $L_{\rm corr}$ works as expected to strongly correlate the predetermined latent components with the objective properties, and with the proposed model, the properties of the generated compounds can be controlled by adding bias to the correlated components.
If the quality of the proposed compound is thus improved, the number of experiments required for the discovery of new materials can be reduced.

Finally, We verified whether we can really reduce the number of experimental trials with the proposed method.
By mimicking the process of discovering new materials with the past experimental data provided by Mitsui Chemicals,
we estimated and compared the number of experiments required to find unlearned but high-performance materials with conventional methods.
First, the outer VAE is trained on 500 thousands open SMILES data set. Next, the inner VAE is trained on the provided real experimental data. The details of this experimental data are omitted due to confidentiality, but there are 100 compounds data. The training is performed with 97 records excluding the top 3 performing records. In other words, the model does not learn information on these high-performing compounds. If the model is able to learn properly even in this situation, it should be able to encode the unknowns into latent variables based on the correlations embedded in the latent space, taking into account the properties of the unknowns. This means that when ranked by the value of $Z^{\rm in}[0]$, the unknown but high performing 
compounds would be ranked in the relatively top group.

\begin{figure}[H]
    \includegraphics[width=0.98\columnwidth]{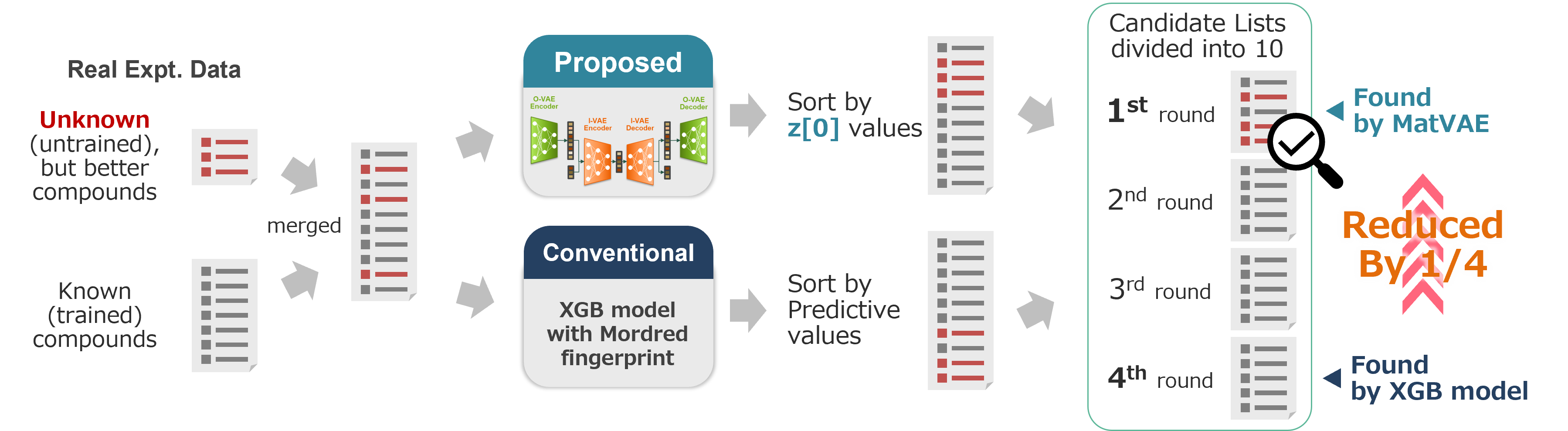}
    \centering
    \caption{
        Schematic illustration of the experimental procedure to estimate whether the number of experiments can be reduced by the proposed method.
        The material development process was simulated using actual material development experimental data.
    } \label{chem_inc_expt}
\end{figure}

After the training, we input all 100 data including unlearned but high-performance compounds data and obtain the latent vectors. Due to the $\mathcal{L}_{\rm corr}$, the $Z^{\rm in}[0]$ surely correlates with the property values, adn we can obtain a list of 100 compounds sorted by $Z^{\rm in}[0]$.
Then, the list is divided into 10 sub lists of 10 items each. The sub lists can be regarded as the candidates lists for a single experimental cycle.
These experiment protocol is described in~\ref{chem_inc_expt}.
We statistically examined the number of cycles where the unlearned but better compounds were found by changing the combinations of top performing compounds.
Experimental results show that in most cases, the XGB model with Mordred Fingerprint requires up to the fourth sub list to discover unlearned but highly-performance compounds, while the proposed method requires only the first sub list, which indicates the MatVAE can reduce the number of experiment trials into about $1/4$.

\newpage

\section{Conclusion} 
\label{sec:conclusion}

In this study, we investigated a deep generative model to shorten the material development time by proposing compounds with high probability of performance improvement as experimental candidates.
This study improved the deep generative model, called MatVAE, which has a nested structure of two VAEs trained with different data set and learning objectives, and verified its performance.
In particular, the loss function $L_{\rm corr}$ was proposed to form the latent space strongly correlated with the target property, which enabled us to generate candidate compounds with desirable properties more directly than the neighbor search in the latent space.
The validation using open data confirmed the validity of using a predefined component $z^{\rm in}[k]$ of the latent vector, which is strongly correlated with the property values to be improved via $L_{\rm corr}$, as a parameter for candidate generation.
In fact, another verification using past experimental data also confirmed that the method can produce high performance compounds, and it was confirmed that the method is effective in reducing the number of experiments required for the discovery of high performance materials by a 1/4 compared to the virtual screening method based on XGB model.

\bibliographystyle{unsrtnat}
\bibliography{matvae_arxiv_20230206}

\end{document}